\documentclass[letterpaper, 10 pt, conference]{ieeeconf}  

\IEEEoverridecommandlockouts                              
 \usepackage{hyperref}
\usepackage[dvipsnames]{xcolor}
\usepackage{graphicx}
\usepackage{amsmath}
\usepackage{hyperref}
\usepackage{etoolbox}
\usepackage{subfigure, subcaption}
\makeatletter
\newcommand{\etal}{\textit{et al.}}
\patchcmd{\@makecaption}
  {\scshape}
  {}
  {}
  {}
\makeatother
\title{\LARGE \bf
Robot Localization and Mapping\\ Final Report\\Sequential Adversarial Learning for\\ Self-Supervised Deep Visual Odometry

}

\author{Akankshya Kar, Sajal Maheshwari, Shamit Lal, Vinay Sameer Raja Kadi \\
\{akankshk,sajalm,shamitl vkadi\} @andrew.cmu.edu}

\newcommand\blfootnote[1]{%
  \begingroup
  \renewcommand\thefootnote{}\footnote{#1}%
  \addtocounter{footnote}{-1}%
  \endgroup
}

\begin{document}
\maketitle
\pagestyle{plain}

\section{INTRODUCTION}

\blfootnote{Our Pytorch implementation: \href{https://bitbucket.org/akankshyakar/slam_project/src/master/slam_project/}{BitBucket repo}}

Visual odometry (VO) and SLAM have been using multi-view geometry via local structure from motion for decades. These methods have a slight disadvantage in challenging scenarios such as low-texture images, dynamic scenarios, etc. Meanwhile, use of deep neural networks to extract high level features is ubiquitous in computer vision. For VO, we can use these deep networks to extract depth and pose estimates using these high level features. The visual odometry task then can be modeled as an image generation task where the pose estimation is the by-product. This can also be achieved in a self-supervised manner, thereby eliminating the data (supervised) intensive nature of training deep neural networks. Although some works tried the similar approach \cite{zhou2017unsupervised}, the depth and pose estimation in the previous works are vague sometimes resulting in accumulation of error (drift) along the trajectory. 

\begin{figure*}[ht]
    \centering
    \includegraphics[width=1.0\linewidth]{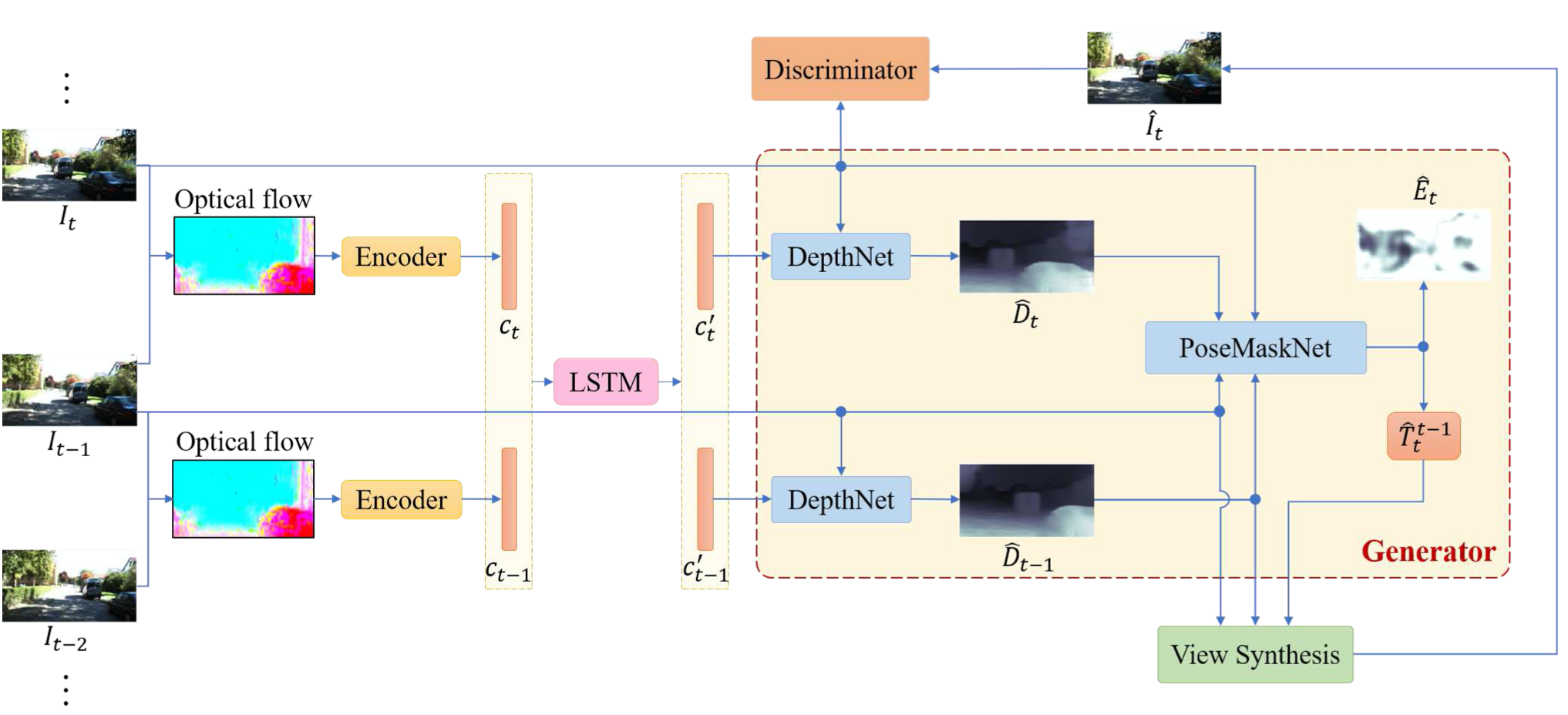}
    \caption{Entire pipeline of the proposed method \cite{li2019sequential}}
    \label{main}
\end{figure*}

The goal of this work is to tackle these limitations of past approaches and to develop a method that can provide better depths and pose estimates.
To address this, a couple of approaches are explored:
\begin{enumerate}
    \item Modeling: Using optical flow and recurrent neural networks (RNN) in order to exploit spatio-temporal correlations which can provide more information to estimate depth. 
    \item Loss function: Generative adversarial network (GAN) \cite{NIPS2014_5423} is deployed to improve the depth estimation (and thereby pose too), as shown in Figure \ref{main}. This additional loss term improves the realism in generated images and reduces artifacts.
\end{enumerate}
The goal of generator is to generate an image by view synthesis which looks exactly like the target image and can be used to fool the discriminator. On the other hand, discriminator in GAN evaluates the generated warped image and tries to distinguish it from the real target image. The eventual goal is to train the generator such that it is hard for the discriminator to distinguish. During inference, only the depth and pose estimators are used to compute the trajectory for VO task. This project would be an implementation of \cite{li2019sequential}. It is built on  the Pytorch implementation of SFMlearner \cite{pytorchsfm}.

However, using GAN to generate the images does not take into account the geometric constraints of the scene. Most recent works based on learning-based techniques have not provided a comparison of their results with those generated using the traditional geometric analysis. We propose to explore this aspect of GAN-based synthesis models and analyze their possible advantages/disadvantages over the traditional methods. We have released our code here: \href{https://bitbucket.org/akankshyakar/slam_project/src/master/slam_project/}{code repo}.
To isolate the performance of each component of our model and validate its contribution to the final goal of high quality depth and pose predictions, we also perform a number of ablation experiments. Unlike \cite{li2019sequential}, we explore different GAN models and algorithms like Wasserstein generative adversarial networks \cite{wgan} and PatchGANs \cite{pix2pix2016}. 

The structure of this report is as follows: In Section II, we will cover the existing literature related to our work. Section III delineates the methodology, different components and the loss functions used. This will be followed by the experiments in Section IV. We discuss the challenges faced in Section V. We finally wrap up the report with Conclusion and Future Work in Section VI.

\section{Literature Review}
\subsection{Pose and Depth estimation}
\label{poseAndDepth}
Pose estimation and visual odometry have traditionally been viewed as a multi-view geometry problems. Initial approaches to solve this problem used graph based optimizations~\cite{mur2015orb,quan1999linear}. These methods typically relied on handcrafted features for finding corresponding points between images and then using these points for subsequent pose estimation~\cite{tareen2018comparative}.

Recently, deep learning based methods have taken over the pose and depth estimation problem as they tend to perform significantly better than the metric-based approaches. Some of the recent works using deep learning for pose estimation are~\cite{zhou2018deeptam, zhou2017unsupervised, wang2017deepvo, li2019sequential, ummenhofer2017demon}. The model proposed by Ummenhofer \etal  in~\cite{ummenhofer2017demon} tries to jointly estimate the pose along with the depth in an end-to-end manner. DeepTAM~\cite{zhou2018deeptam}, on the other hand, uses two different networks for pose estimation and depth estimation. However, these methods require supervision for training. The models, therefore, can only be trained on synthetic data as supervised data generation for pose estimation and depth estimation in the wild is both expensive and heavily prone to inaccuracies.

In order to overcome this problem, there have been numerous self-supervised methods that have recently been proposed. These methods include SFMLearner by Zhou \etal~\cite{zhou2017unsupervised}. Since the current work that we present in the report~\cite{li2019sequential} is built over this method, we explain the SFMLearner model in detail.

The SFMLearner model utilizes the correlation between depth and pose to learn both entities jointly. The model makes use of a photometric and geometric loss to model the constraints between the source and the target pixel in an image as a function of its depth, pose change and the camera intrinsic. Self-supervision is used to train these models by using novel view synthesis. View synthesis is predicting what a scene will look like from a different view. This is done by unprojecting the image in source view using camera intrinsics and depth image, warping it to the pose of the target view using the source to target relative pose transformation, and finally projecting it to the  pixel space. Differential modules are used for view synthesis task in order to take advantage of gradient based training ~\cite{cheng2018geometry, tung20203d, lal2021coconets, xian2021hyperdynamics, prabhudesai2020disentangling, prabhudesai20203d}. Authors show the performance of their method on KITTI~\cite{Geiger2012CVPR} and Make3D datasets \cite{saxena2008make3d}. The results show that the single view depth prediction gives results equivalent to the supervised learning approaches on KITTI dataset. Ablation studies showed that explainability network only offered a small boost in performance, probably because most scenes in KITTI are static. The results on Make3D dataset shows the generalization ability of the depth prediction model. For pose estimation network, authors use the KITTI odometry split, and show that their network performs comparable to ORB-SLAM.

This work was further extended in~\cite{li2019sequential}, by incorporating sequential learning as part of the VO framework, estimating depth conditioned on multiple previous frames, in contrast to the single view depth prediction. This work is explained in more detail in Section~\ref{sec:method}.

\subsection{Generative Adversarial learning}
A majority of the self-supervised approaches methods mentioned in Section~\ref{poseAndDepth} need to generate photo-realistic images as part of the target image. This is done using Generative Adversarial Networks or GANs as they are commonly referred to. These were first proposed by Goodfellow et al. in~\cite{NIPS2014_5423} to try to generate realistic images from a latent space by trying to model the loss as a mini-max function between the two components (Generator and Discriminator) of the architecture. GANs have been a topic of extensive study for the past few years and many improvements and variants over the original GAN architecture have been suggested~\cite{wgan, isola2017image, radford2015unsupervised}. GANs have also been used with great success in a number of image editing applications ~\cite{zhang2016colorful, jing2019neural, gatys2016image, frid2018gan, yu2019free, yeh2017semantic, lal2017automatic, ledig2017photo, karras2017progressive}. In this work, we have used two variants as part of the discriminator in the adversarial training. These are the PatchGAN architecture proposed in~\cite{isola2017image} and the WGAN method in~\cite{wgan}. The details for the same are provided in Section~\ref{sec:method}.

\section{Methodology}
\label{sec:method}
Our network used for estimation of depth and pose in an self-supervised manner consists of the following 4 components:
\subsection{Sequence representation}

Since there is lot of redundant information in the images for estimating pose and depth, the optical flow between consecutively sampled frames is extracted and computed into a 128 dimensional vector, using and encoder. Let us denote it by $c_t$. These $c_t$ vectors are passed to LSTM to leverage the correlations in sequence. The output of LSTM at each time stamp, denoted by ${c_t}'$, is then passed into next module. This ${c_t}'$ is expected to contain both sequence context and optical flow information. This way we can encode the spatio-temporal aspect in a video sequence.
\subsection{Depth estimation}
Although ${c_t}'$ contains much information in the form of correlations, compared to just a single image, we pass the current image also as an input to depth prediction network at pixel level. In earlier works, they use a encoder-decoder architecture for estimating depth using a single image. This is an ill posed problem. But here, instead of just estimating depth using a single image, we pass the encoded LSTM output at the current time to estimate the depth using an up-sampling decoder. DispNet \cite{mayer2016large}, an architecture which used encoder-decoder design with skip connections and multi-scale side predictions, is used for our Depth prediction.
\subsection{Pose estimation}
Once we have the depth estimation, we then use it along with the respective images to regress for the pose prediction. Pose network predicts the transformation parameters (Rotation (R) and translation (T)) between a source frame and a target frame. Considering only one pair of source and target frames, these parameters are used to calculate the corresponding locations in source for each location in target frame, i.e take a grid/matrix with each entry containing the pixel location in target frame and multiply with intrinsics, depth estimation and transformation parameters to get the corresponding pixel location in source frame from which we have to sample. Now, we use a bi-linear sampling kernel to create a mask which is multiplied with the source image (pixel values) to get the predicted view. This is the same procedure followed in \cite{jaderberg2015spatial}. Thus the output of PoseNet is used to warp the source image to get the target image. Now, whatever loss we compute on the predicted view (such as reconstruction loss) can be back-propagated to train the PoseNet parameters as the sampling procedure described above is nothing but a series of matrix multiplications.
For robustness, we calculate explainability mask to apply loss only on non-occluded pixels. 
\subsection{Adversarial training}
As photo-metric loss can get stuck at a local minima due to occlusions, dynamic objects and texture-less regions, using an adversarial training step can lead to a better depth and pose prediction (better learning of parameters). Since, we have access to the original frame (the current time step frame) that we are trying to predict (using view synthesis) using depth and pose inputs, we can use a discriminator to evaluate the quality of prediction and back-propagate this loss to learn depth and pose network parameters.
\\
\subsection{Loss Function}
We now delineate the losses we use to train our model.
\paragraph{\textbf{Appearance Loss}} Appearance loss is given by:
\begin{equation}
    \mathcal{L}_{ap} = \mathcal{L}_{reg}(\hat{M}) + (1-\alpha)\mathcal{L}_{pho} + \frac{1}{N} \mathcal{L}_{ssim}
\end{equation}
where N is the minibatch size and $\alpha$ is set to 0.85.
\\
$\hat{M}$ is the per-pixel explainability mask which the model predicts and it is a belief on whether a pixel is occluded in target frame or not. $\mathcal{L}_{reg}$ therefore helps in preventing the trivial solution where $\hat{M}$ will just become 0, thereby making $\mathcal{L}_{pho} = 0$ too. It does so by penalizing the network for predicting values smaller than 0. It, therefore, works as a regularizer.
\\
$\mathcal{L}_{pho}$ is the loss over how well the predicted image matches the target image. It is given by:
\begin{equation}
    \mathcal{L}_{pho} = \sum_{<I_1,...,I_N>} \sum_{p} \hat{M}_{t}(p) ||\hat{I}_t(p) - I_t(p) ||_{1}
\end{equation}
where $<I_1,...,I_N>$ are the sequence frames, p is the number of pixels in the image, $\hat{I}$ represents the predicted image and $I$ is the target image. 
\\
Finally, $\mathcal{L}_{ssim}$ \cite{ssim} is the loss over perceived quality of the image and is given by:
\begin{equation}
    \mathcal{L}_{ssim} = \sum_{x,y} \alpha \frac{1-SSIM(\hat{I}(x,y), I(x,y))}{2}
\end{equation}
We use a filter size of $10 \times 10$ for SSIM. 


\paragraph{\textbf{Trajectory Loss}} The trajectory loss imposes a geometry constraint over the poses predicted by the network. Suppose A, B, C, D are 4 frames of a sequence. Let $A \rightarrow B$ be the 6-DOF relative transformation that transforms from state A to state B. Given this, the trajectory loss penalizes the network when the relative transformation $A \rightarrow D$ predicted by the network deviates from the $A \rightarrow D$ transformation computed by combining the individual relative transforms $A \rightarrow B$, $B \rightarrow C$, $C \rightarrow D$.
The trajectory loss is given as:
\begin{equation}
    \mathcal{L}_{tc} = \frac{1}{N}\sum_{i=1}^{N}\sum_{t\in[2,4,8]}|| \hat{p}_{d_i}^{i+t} - \hat{p}_{r_i}^{i+t}||_{1}
\end{equation}
where $\hat{p}_{d_i}^{i+t}$ is the 6-DOF pose estimated directly for $A \rightarrow D$, and $\hat{p}_{r_i}^{i+t}$ is the pose computed by concatenating individual relative poses.
We apply trajectory loss for intervals of size 2, 4, and 8.

\paragraph{\textbf{Depth Regularization:}}
Depth regularization enforces local smoothness and discontinuity in the predicted depth. Depth discontinuity usually happens where there are strong image gradients. This loss is given by:

\begin{align}
\begin{split}
    \mathcal{L}_{smo} &= \frac{1}{N}\sum_{x,y}||\nabla_x \hat{D}(x,y)||e^{-||\nabla_x I(x,y)||}  + 
    \\
    &||\nabla_y\hat{D}(x,y)||e^{-||\nabla_y I(x,y)||}
\end{split}
\end{align}

\paragraph{\textbf{GAN Loss:}} GAN loss helps ensure that the image generated by the model will look real and will not have weird artifacts that are not present in real images. It is given  by:

\begin{align}
\begin{split}
    \mathcal{L}_{GAN} &= \min_G \max_D V(G,D)\\&= E_{I_t \sim p_{real}}[log(D(I_t|I_t))] \\&+ 
    E_{c_{t-1}^{'}, c_t^{'} \sim p_{code}}[1 - log(D(\hat{I}_t|I_t))]
\end{split}
\end{align}

The final loss is the weighted sum of the losses mentioned above and is given by:
\begin{equation}
    \mathcal{L}_{final} = \lambda_a\mathcal{L}_{ap} + \lambda_s\mathcal{L}_{smo} + \lambda_t\mathcal{L}_{TC} + \lambda_g\mathcal{L}_{GAN}
\end{equation}
\section{Experiments}
\subsection{Dataset}
The model is evaluated on the KITTI dataset ~\cite{Geiger2012CVPR} in our experiments. The dataset contains videos of outdoor sequences. The depth dataset contains pairs of RGB images along with the depth maps for over 90,000 images. The visual odometry dataset in the benchmark consists of 22 sequences with half the sequences used for training with ground truth trajectories available.
\subsection{Implementation Details}
\subsubsection{Optical Flow}
\begin{figure}
\centering
\begin{subfigure}{}
    \includegraphics[width=115px]{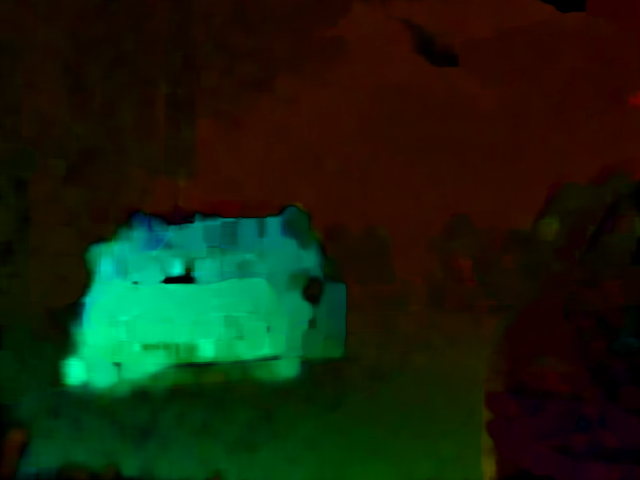}
\end{subfigure}
\begin{subfigure}{}
    \includegraphics[width=115px]{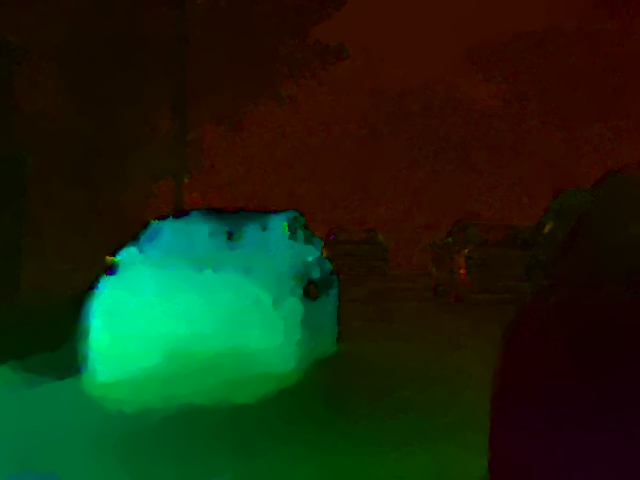}
\end{subfigure}
    \caption{Comparison of Farneback and Ce Liu's Coarse2Fine optical flow methods }
    \label{OF}
\end{figure}

To obtain the optical flow, we evaluated two methods, namely, Farneback method ~\cite{fback} and Coarse2Fine optical flow method~\cite{celiu}. Among these, we noticed that although the latter has better quality compared to the former, it is much slower compared to the former. So, we finalized on Farneback method based on its much faster speed (10X) and its output (along with image input) seemed good enough to help depth estimation even though it is not as perfect as Coarse2Fine method. The preliminary results of our experiment with optical flow can be seen in Figure \ref{OF}
\subsubsection{Encoder}
The optical flow images obtained from above are passed to an Encoder network to obtain the optical flow encoding. The Encoder network consists of 6 down-sampling convolution layers followed by an adaptive 2D average pooling layer. Each down-sampling convolution layer in turn consists of two convolution layers with stride 2 for the first convolution layer. Batch normalization is used on the output of each of these convolution layers. The number of channels in the 6 down-sampling layers are $\{16, 32, 64, 128, 128, 128\}$ respectively and the kernel sizes are $\{7,5,3,3,3,3\}$ respectively (kernel size is (i,i) for i in the mentioned list). The weights are initialized using xavier\_uniform initialization whereas the biases are initialized to 0.
\subsubsection{LSTM}
The optical flow encodings are further passed to LSTM to leverage the sequential information present in videos. We used a 1 layer LSTM network with a hidden dimension of 128 and took the h state at every time as the improved optical flow encodings.
\\
\subsubsection{Disparity Network}
The optical flow encodings and images are fed to a disparity network which predicts the disparity(=1/depth). Similar to the previous works~\cite{li2019sequential}, \cite{zhou2017unsupervised}, we predict depth at 4 different scales. First the image is passed through a series of 7 downsample convolution layers, with each downsampling conv layer same as mentioned earlier. The optical flow encoding is added to the lowest level (output of $7^{th}$ layer) and the resulting tensor is passed to an unsampling convolution layer to upsample using transposed convolutions. The upsampled tensor is concatenated with downsampled tensor at the higher level and passed to an upsampling convolution layer. This is repeated for all the 7 levels and resulting tensors at first 4 levels form our disparity predictions after doing a bilinear interpolation followed by another conv layer where we use Sigmoid activation function and a scaling($\alpha$) and shift($\beta$) hyperparameters to constrain the disparity values to a certain range. In our experiments, $\alpha$ and $\beta$ are set to 10 and 0.01 respectively. The weights are initialized using xavier\_uniform initialization whereas the biases are initialized to 0.
\\
\subsubsection{Pose Network}
The output depth maps are fed to the a pose estimation network along with the sequence image at the current and the previous time step. The pose estimation network consists of seven convolution layers with the number of activation maps at each convolution output equal to $\{16, 32, 64, 128, 256, 256, 256\}$ respectively. The kernel sizes for the first two layers is equal to 7 and 5 respectively. The rest of the convolution filters have a kernel size of 3. We use ReLU as the non-linear activation function. The output of the convolution layer is fed to two sub-modules. These sub-modules consist of a decoder architecture similar to the Flownet architecture~\cite{dosovitskiy2015flownet} which outputs an explainability mask to discard the occluding/moving points which might be producing erroneous correspondences for pose estimation. Since the range of the explainability mask is [0,1], we use a Sigmoid activation at the at the output of the decoder layers to generate the explainability mask(at all scales). Similar to depth network, we predict the explainability mask at 4 different scales. The second sub-module consists of a single convolution layer with 6 output channels corresponding to the 6 DOF for pose estimations. This output represents the relative pose change between the source and the target image, and is used for the camera trajectory estimation. The weights are initialized using xavier\_uniform initialization whereas the biases are initialized to 0. \\
\subsubsection{GAN}
We experimented with multiple discriminator architectures for the adversarial training. We used WGAN and PatchGAN to study the effects of using a global and local discriminator. \\
The authors in~\cite{li2019sequential} have suggested WGAN\cite{wgan} with the same architecture as the encoding part of the DispNet. SGD optimiser is used and WGAN. The loss function here is (Average critic loss of fake images) - (Average critic loss of real images). 
The PatchGAN architecture used here is the architecture proposed in~\cite{isola2017image} and since then has been widely used for image-image translation problems. The discriminator tries to classify if a $N \times N$ patch is real or fake. The final output of the discriminator is the average output of the discriminator results over all the input patches. The implementation has been borrowed from \cite{ganimp}
\\
The weight for each of the loss terms in the total loss (for all ablation experiments) can be seen in README section in the code link provided. The default values of the weights are 0.75, 0.1, 0.14 and 0.01 for appearance, smoothness, trajectory and GAN loss respectively. We use Adam optimizer in our experiments.

\subsection{Evaluation metrics}
For evaluating the predicted depth images, we use 6 different metrics as suggested in \cite{li2019sequential}. The Absolute difference metric calculates the mean of absolute value of pixel depth value differences between ground-truth and predicted depth images. The Absolute relative metric differs from the above by normalizing the depth value differences using the original(ground-truth) depth values. The Square relative metric differs from Absolute relative metric by using square instead of absolute values while calculating the differences.

For the rest of the metrics, we first introduce the term $\delta$ given by $ \delta = max\left ( \frac{\text{predicted depth}}{\text{ground truth depth}}, \frac{\text{ground truth depth}}{\text{predicted depth}} \right )$. We then take three thresholds 1.25, $1.25^2$, $1.25^3$ and for each of these values, we calculate how many percentage of the pixels satisfy the condition $\delta<threshold$. Other than these 6 metrics, we use the standard ATE(Absolute Trajectory Error) metric for evaluating the trajectory and thereby poses.
\subsection{Results}
In this section, we present the experimental results of the current pipeline benefiting from the additional information in form of the sequential inputs over the erstwhile state-of-the-art SFMLearner pipeline~\cite{zhou2017unsupervised}. We also try to study the effects of different components of the pipeline used by performing extensive ablation studies. The depth estimations are evaluated using the metrics described above. The pose estimations are measured using the Absolute Trajectory Error (ATE) for both translation and rotation.

\subsubsection{Ablation Studies}
In this section, we show the results of ablation studies of our model and visualize the predicted depths and loss curves for each ablation experiment. All the experiments use a sequence of 15 images of which we calculate the optical flow. Now these optical flow may contain redundant information and we encode these to get get a code. This code is the input to all our ablations.

\begin{figure}[h]
    \centering
    \includegraphics[width=1.0\linewidth]{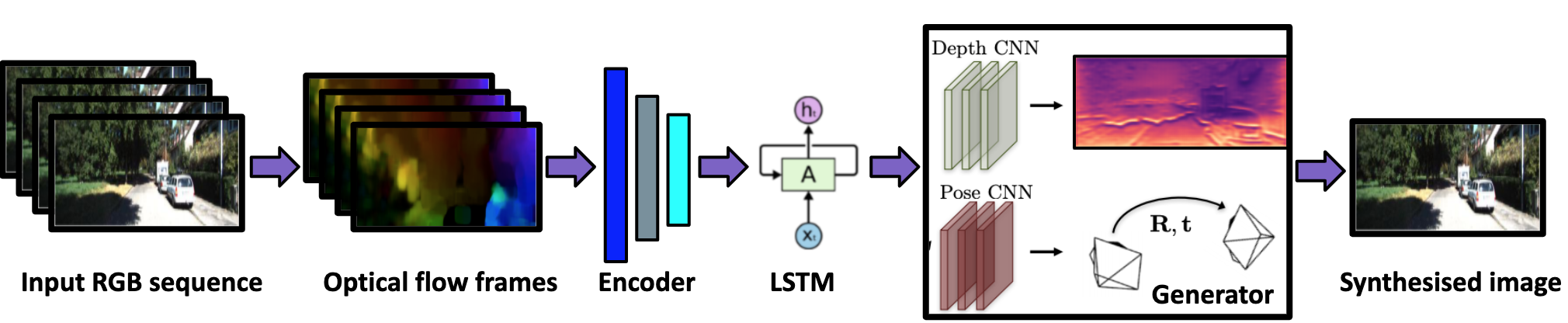}
    \includegraphics[width=1.0\linewidth]{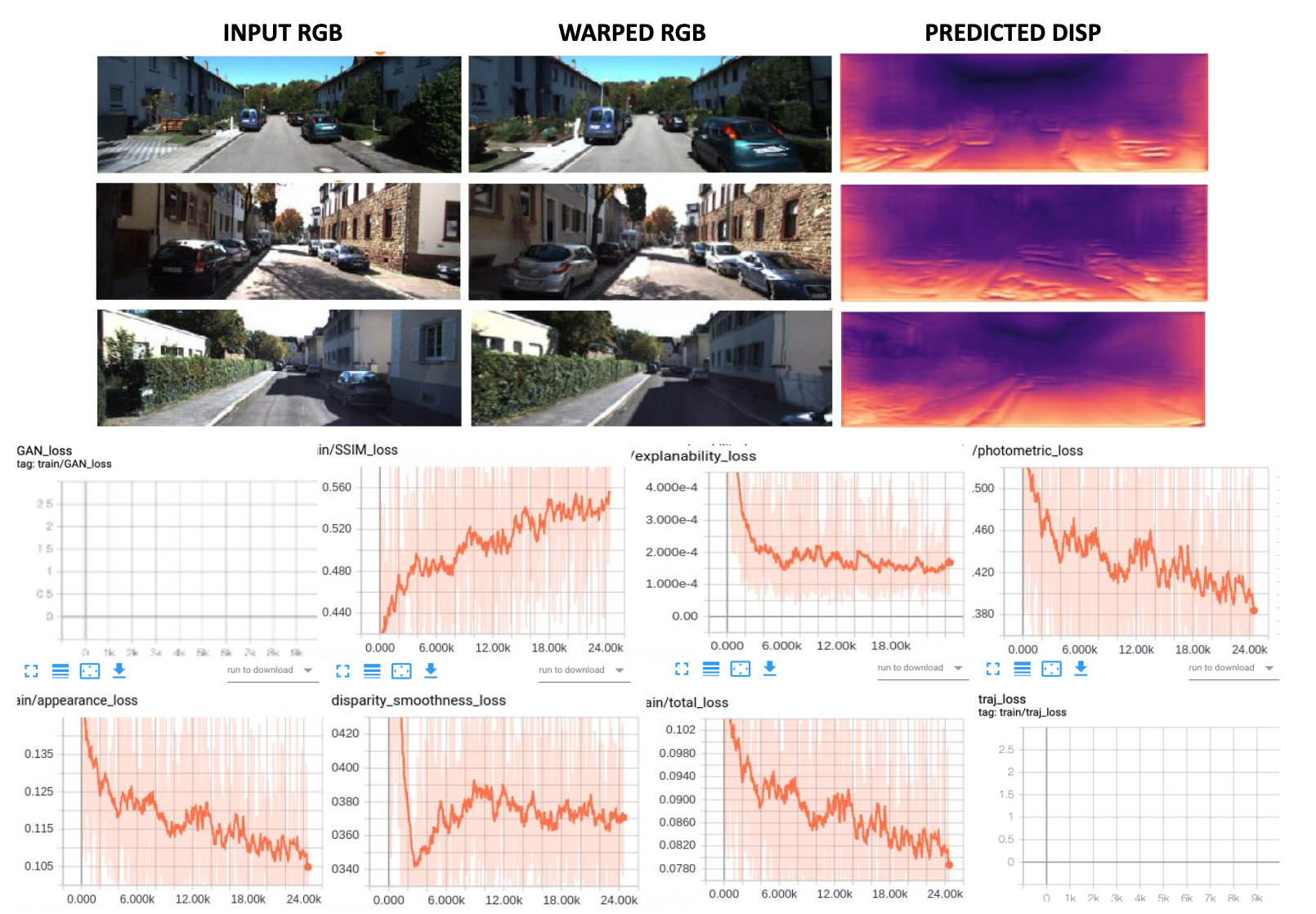}
    \caption{Qualitative results and loss curves for SFMLearner  + OpticalFlowCode + LSTM ablation.}
    \label{fig:galaxy}
    \label{bcl}
\end{figure}

\paragraph{\textbf{SFMLearner  + OpticalFlowCode + LSTM}}
In this ablation experiment, we augment the SFMLearner model with LSTM and optical flow computation. The significance of this experiment is we want to leverage the sequential information in a video to predict better pose and depth. This will result in better warped image. Simple photo-metric loss is used to check view synthesis. The predicted depths and warped RGBs are shown in Figure \ref{bcl}.

\begin{table*}
\begin{center}
\begin{tabular}{|l|c|c|c|c|c|c|}
\hline
{} & {Abs.Diff}&{Abs.rel} & {Sq.rel} & {$\delta < 1.25$} & {$\delta < {1.25}^{2}$} & {$\delta < {1.25}^{3}$}  \\ \hline 
 {SfmLearner}& {5.12} & {0.443} & {4.24} & {0.517} & {0.765} & {0.811} \\ \hline 
 {SFMLearner + OpticalFlowCode + LSTM} & {4.74} & {0.243} & {2.27} & {0.518} &{0.761} & {0.857}\\\hline
 {SFMLearner  + OpticalFlowCode + LSTM + WGAN}& {5.08} & {0.259} & {2.7} & {0.519} & {0.736} & {0.83}\\\hline
 {SFMLearner  + OpticalFlowCode + LSTM + PatchGAN}& {\textbf{4.50}} & {\textbf{0.237}} & {2.25} & {\textbf{0.56}} & {\textbf{0.775}} & {\textbf{0.861}}\\\hline
 {SFMLearner  + OpticalFlowCode + WGAN}& {5.361} & {0.266} & {3.0} & {0.507} & {0.717} & {0.815}\\\hline
 {SFMLearner + OpticalFlowCode + LSTM + WGAN + Trajectory Loss }& {4.089} & {0.295} & {\textbf{2.18}} & {0.461} & {0.706} & {0.821}\\\hline
 {SFMLearner + OpticalFlowCode + LSTM + PatchGAN + Trajectory Loss }& {4.874} & {0.248} & {2.55} & {0.539} & {0.745} & {0.837}\\\hline
\end{tabular}
\end{center}
\caption{Comparison of different approaches on KITTI dataset for depth estimation model}
\label{tab:depth}
\end{table*}

\begin{figure}[h!]
    \centering
    \includegraphics[width=1.0\linewidth]{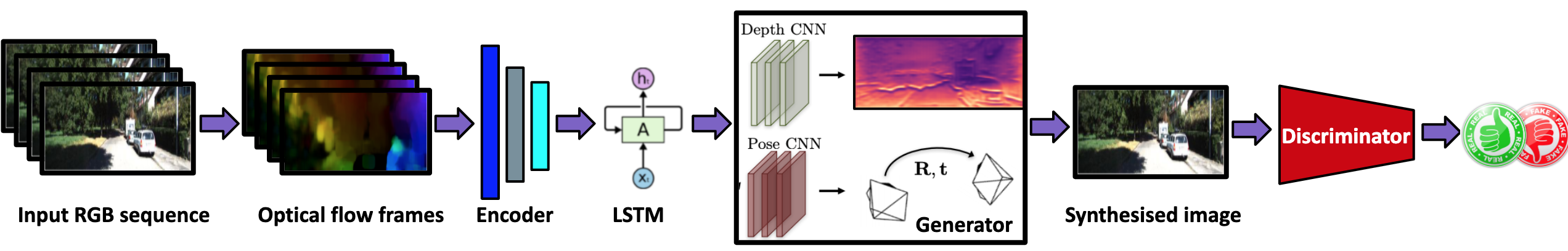}
    \includegraphics[width=1.0\linewidth]{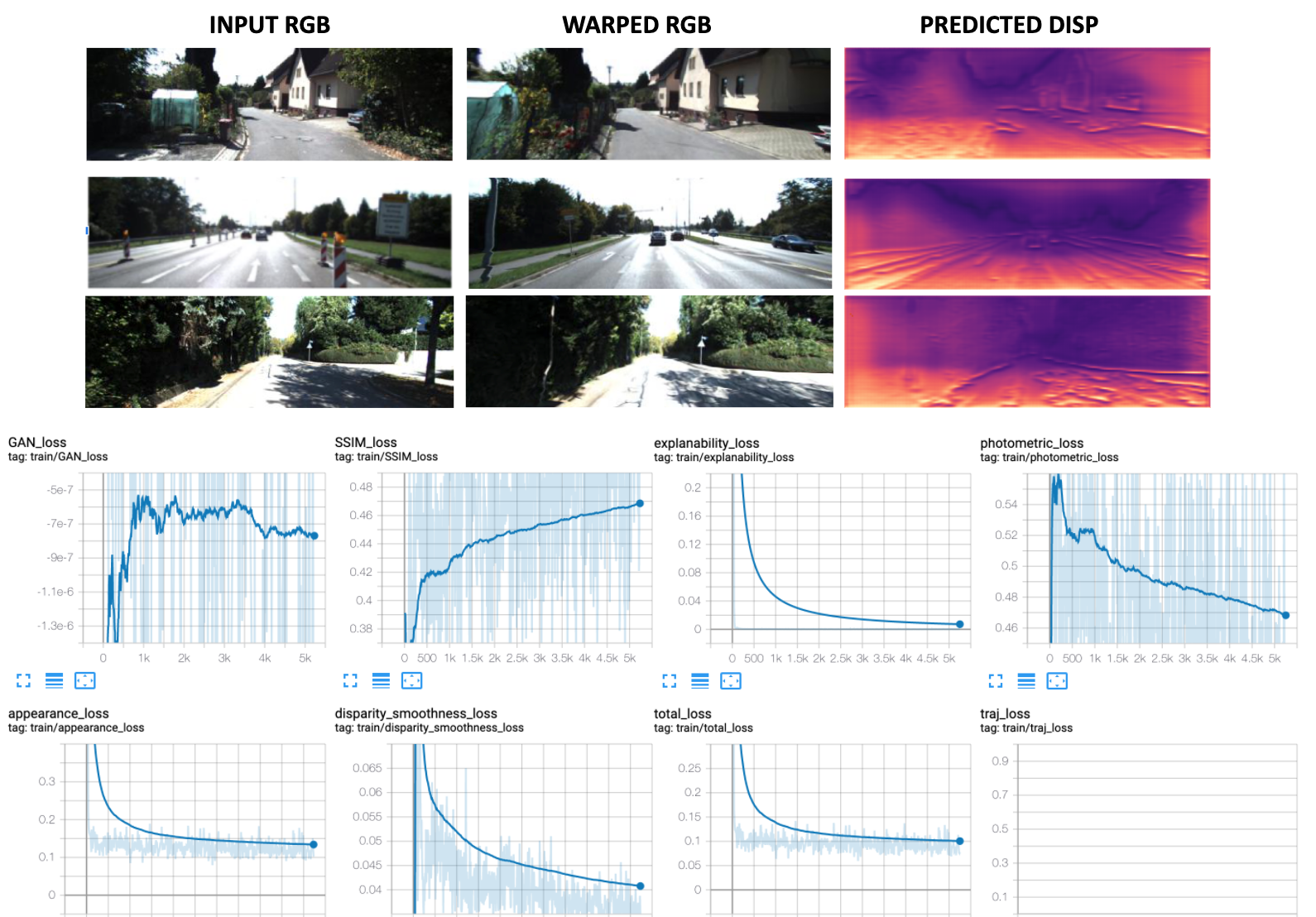}
    \caption{Qualitative results and loss curves for SFMLearner  + OpticalFlowCode + LSTM + WGAN ablation.}
    \label{fig:galaxy}
    \label{bclwgan}
\end{figure}

\paragraph{\textbf{SFMLearner  + OpticalFlowCode + LSTM + WGAN}}
In this experiment, we add the GAN loss in the model. This is introduced by adding a discriminator network. The task of discriminator is to find the fake ones. The warped image is given as input. We use WGAN for this ablation. This loss is the same as used by authors in \cite{li2019sequential}. This loss helps in predicting the real or fake warped image without collapsing. Results are shown in Figure \ref{bclwgan}.

\begin{figure}[h!]
    \centering
        \includegraphics[width=1.0\linewidth]{images/BCLG_diag.png}
    \includegraphics[width=1.0\linewidth]{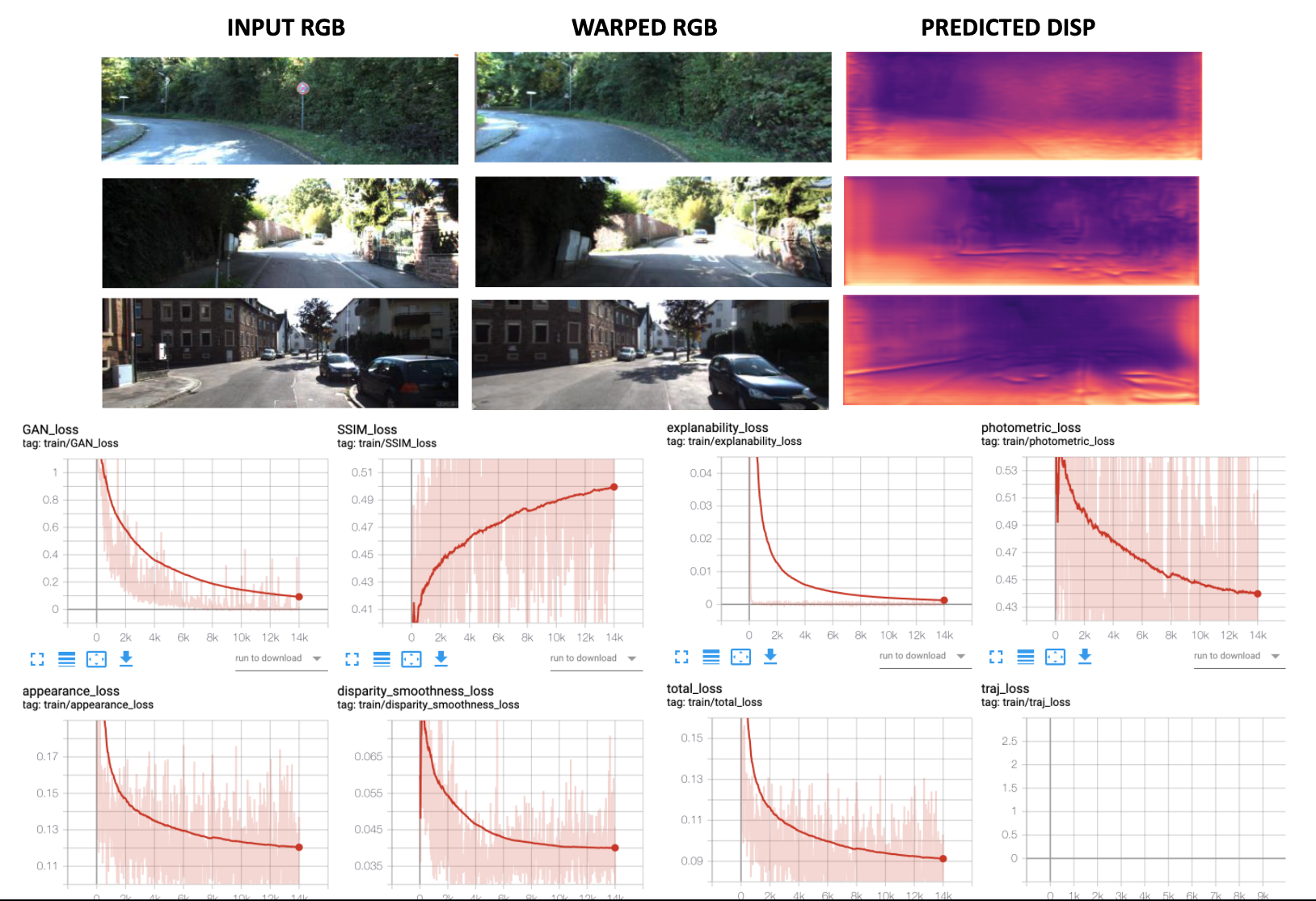}
    \caption{Qualitative results and loss curves for SFMLearner  + OpticalFlowCode + LSTM + PatchGAN ablation.}
    \label{fig:galaxy}
    \label{bclpgan}
\end{figure}

\paragraph{\textbf{SFMLearner  + OpticalFlowCode + LSTM + PatchGAN}}
Here we perform the ablation by replacing WGAN in previous experiment with PatchGAN. PatchGan is one of the most popular GAN networks. here we use the discriminator from this architecture and use MSE loss. The idea is that this will be able to judiciously distinguish real from fake warps and take care of visual artifacts.  The results are shown in Figure \ref{bclpgan}

\begin{figure}[h!]
    \centering
     \includegraphics[width=1.0\linewidth]{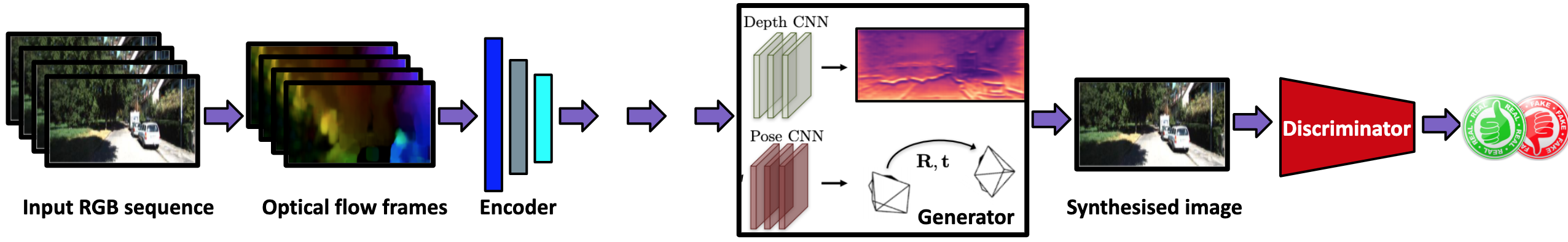}
    \includegraphics[width=1.0\linewidth]{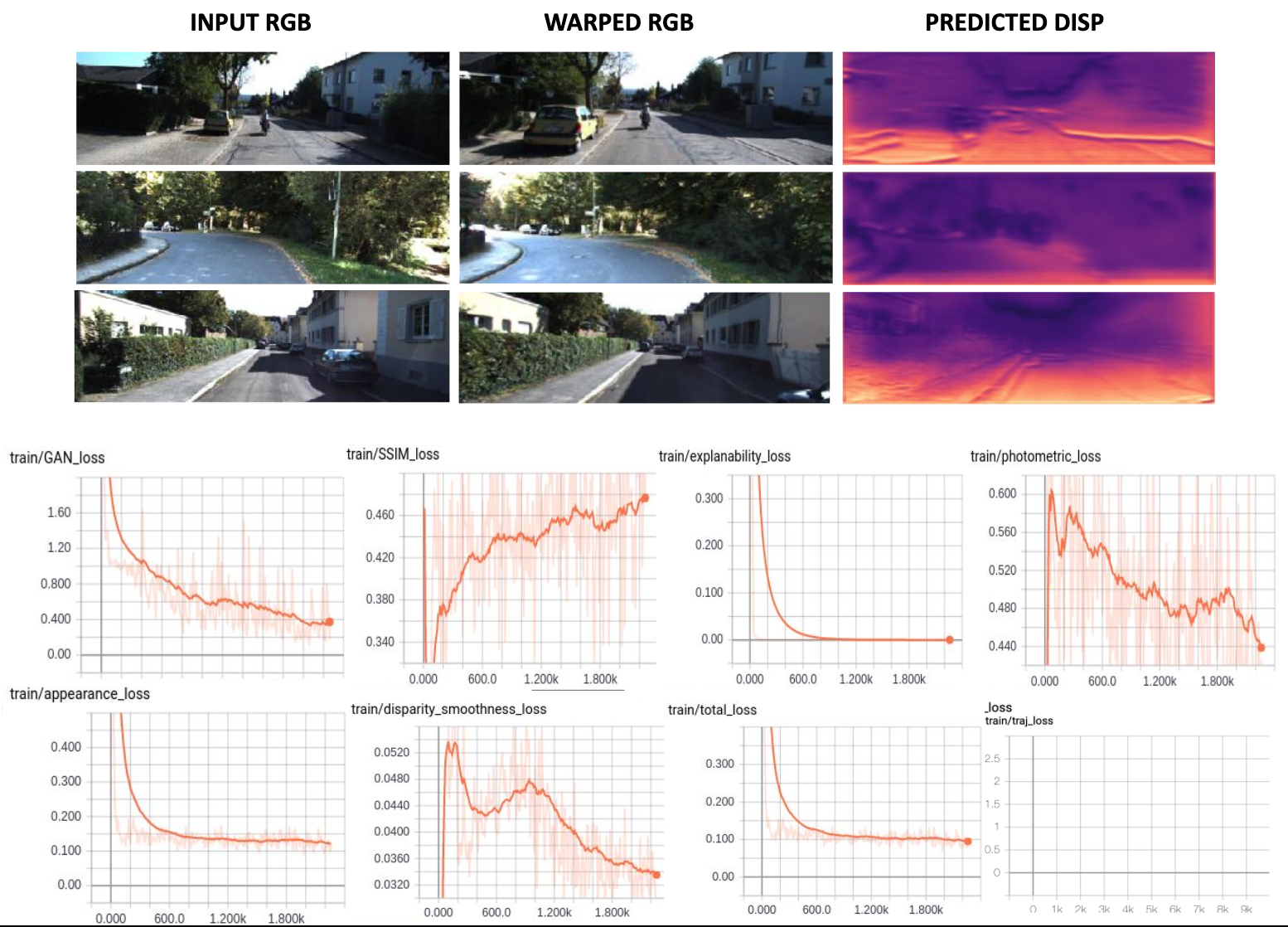}
    \caption{Qualitative results and loss curves for SFMLearner  + OpticalFlowCode + PatchGAN ablation.}
    \label{fig:galaxy}
    \label{bcpgan}
\end{figure}

\paragraph{\textbf{SFMLearner  + OpticalFlowCode + PatchGAN}}
Now, we remove LSTM from our previous experiment model. The significance of this experiment is that we want to see if the recurrence relation is actually captured by LSTM, or Optical flow does a good enough job to encode that information.  The results for this ablation are shown in Figure \ref{bcpgan}.

\begin{figure}[ht]
    \centering
    \includegraphics[width=1.0\linewidth]{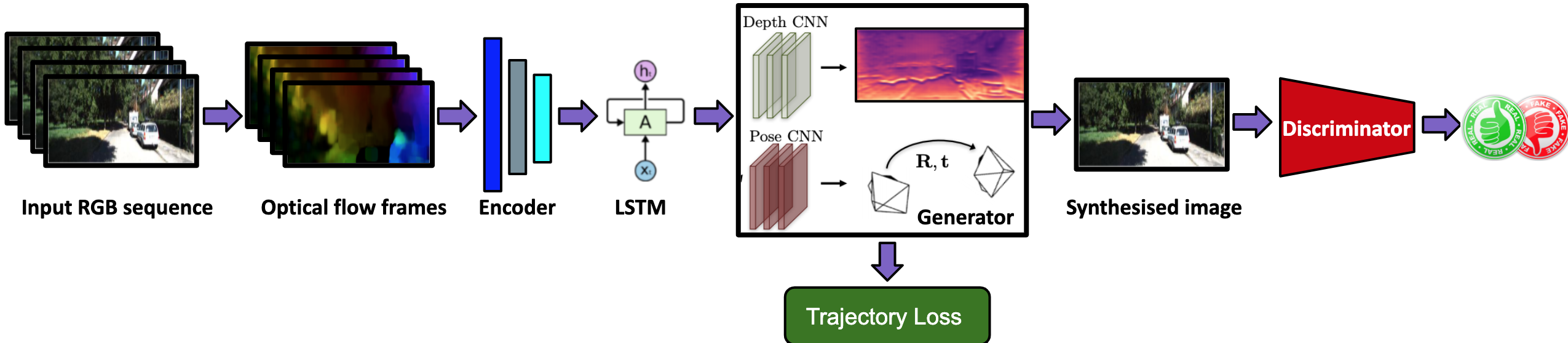}
    \centering
    \includegraphics[width=1.0\linewidth]{images/BCPGAN.png}
    \caption{Qualitative results and loss curves for SFMLearner  + OpticalFlowCode + LSTM + PatchGAN + Trajectory Loss ablation.}
    \label{fig:galaxy}
    \label{all}
\end{figure}

\paragraph{\textbf{SFMLearner + OpticalFlowCode + LSTM + PatchGAN + Trajectory Loss}}
In this final ablation, we include all the components in our model, including the trajectory constraint. The trajectory loss adds geometric consistency constraint between relative poses predicted for different frames. This is especially needed as the other losses  do not care about this. As all our objects in the dataset are rigid bodies, we can use this as our final experiment. The results  are in Figure \ref{all}.
\subsubsection{Depth estimation}
The quantitative results for the depth estimation module can be seen in Table~\ref{tab:depth}. For first 3 lower values are better and last 3 higher values are better. It can be seen our model outperforms the model proposed in~\cite{zhou2017unsupervised}. We also present the results of our model along with various versions of our implementation. We observe that the PatchGAN architecture for the discriminator produces better results than the WGAN model proposed in ~\cite{li2019sequential}. A possible reason for this might be that since the PatchGAN model tries to discriminate real and fake images at a local scale, it might lead to images that are more consistent. \\
\subsubsection{Pose estimation}
The quantitative results for the pose estimation module can be seen in Table~\ref{tab:pose}.
The results demonstrate that this architecture outperforms the SFMLearner architecture by a significant margin. It can be seen here that although the effect of trajectory loss was not quite seen in the depth estimation results, it is apparent that trajectory loss is helpful for correct pose estimation. We see that removal of LSTM has the largest impact on the results, and PatchGAN performs better than WGAN, which is in line with the observations above. 
\begin{table}
\begin{center}
\begin{tabular}{|l|c|c|c|}
\hline
{} & {Seq. 9} & {Seq. 10} \\ \hline 
 {SFMLearner}& {0.006} & {0.0054} \\ \hline 
 {B + Optical Flow + LSTM}& {0.0048} & {0.0046} \\\hline
 {B + Optical FLow + LSTM + WGAN}& {0.0045} & {0.0044} \\\hline
 {B + Optical Flow + LSTM + PatchGAN}& {0.0041} & {0.0037} \\\hline
 {B + Optical flow + WGAN}& {0.0052} & {0.0050} \\\hline
 {B + Optical Flow + LSTM + WGAN + TC}& {0.0039} & {0.0037}\\\hline
 {B + Optical Flow + LSTM + PatchGAN + TC}& {\textbf{0.0034}} & {\textbf{0.0031}}\\\hline
\end{tabular}
\end{center}
\caption{\raggedleft{Comparison of different approaches on KITTI visual odometry dataset for pose estimation model. Here, '+' means that the model consists of the components present in the operands on which '+' operator is applied. 'B' refers to the baseline SFMLearner and 'TC' refers to the trajectory constraint.}}
\label{tab:pose}
\vspace{-1cm}
\end{table}
\section{Challenges}
Some of the challenges we faced during this project are:
\begin{enumerate}
    \item Optical flow computation: Optical flow can either be computed at runtime or can be pre-computed for the entire dataset. Pre-computing optical flow consumes a lot of disk space for KITTI dataset but saves time during training. Computing optical flow at runtime, even though cheap on storage, increases the training time, based on the algorithm being used.
    \item Generative adversarial training: Training GANs come with all the problems related to it. Some of these issues are mode collapse, non-convergence, diminished gradient, etc. In WGAN we noticed the values fluctuating a lot and not having proper convergence. Perhaps that is why the results from WGAN were not the best. In PatchGAN the values were diminishing but at a slower rate. The training time was almost 1.5 times the WGAN training time.
    \item LSTMs: Training LSTMs is also challenging due to problems like vanishing/exploding gradients, unstable training, etc. We did not keep track of the LSTM values, but experiments without LSTM gave poorer performance.
    \item Hyper-parameters: There were lot hyper-parameters to be set in this as there were many loss functions. The values given by the authors in the \cite{li2019sequential} did not work for our implementation, especially for depth regularisation and trajectory loss. The weight values given by the authors for trajectory loss would overpower all other losses and make PoseNet converge to the trivial solution of predicting no change in pose. After that the network would never recover. We tried lot of values in the limited compute capability we had. The hyper-parameter selection especially the weights of different loss functions could be explored further and may give better results.
    \item Code base: There were many sub-modules in the code and they had to be modular enough to plug and play for different ablation experiments. Maintaining that and conceptualising the entire pipeline was difficult. 
    \item GPU availability: We all had limited compute available to train all the networks for multiple iterations.
    \item Instability in training: We had Encoder, LSTM, Dispnet, PoseNet and Discriminator network being trained simultaneously. There was a lot of instability and they may have gone through collaborative self-supervision, where in they would have corrected each other's mistakes to some extent.
    \item SSIM loss: SSIM ranges from $[-1,1]$ with 1 being equal images. The equation in~\cite{li2019sequential} suggested the SSIM value as a part of the appearance loss. However, intuitively, it made more sense to use (1 - SSIM) as a part of the loss function. We referred multiple papers to confirm the correctness of our change. The logic is that we want to penalise dissimilar images and increase loss when the warped image is not similar to GT image. 
\end{enumerate}
\section{Conclusion and Future Work}
In this project we have tried to implement \cite{li2019sequential} paper. We try to leverage temporal consistency in video sequence to solve Visual odometry as a View synthesis problem. We take up sequence of input RGB frames, find optical flow between consecutive frames then pass them through an encoder to get the code. The code is passed to an LSTM which tries to incorporate long term and short term information in the code. This information is passed to the Generator module which consists of a DisparityNet and PoseNet. These 2 are used to generate a warped target image. This target image is then judged for being real or fake by our Discriminator. In order to add some novelty we experimented with PatchGAN as well as WGAN, and found that PatchGAN actually performed better than WGAN for our experiments, the reason for which we have already provided above. This project had multiple loss functions and outputs which harmoniously blend together to solve Visual Odometry task. It took us a long time to get each module working independently and then together with the other modules. In the future we would like to do a more exhaustive hyperparameter search.  We would also like to explore more losses that can add better constraints on the overall system. We have thought of another loss, i.e. cycle consistency loss which can be used in this. Instead of just image generation we can do an inverse warp to get the original image back and they should be same. This might make the system more robust. We also did not experiment with various architectures for PoseNet and DispNet. We mainly tried to use the architectures borrowed from SFMLearner. We would also like to explore some other newer Depth predicting architectures as that might increase the performance further. We would also want to train on Cityscapes dataset \cite{cordts2016cityscapes} and compare the result for that as well.
As a concluding remark, it was a great project to work on and it gave us exposure in multiple areas in Vision and SLAM.

\bibliographystyle{IEEEbib}
\bibliography{references}
\end{document}